# Moderate Environmental Variation Promotes the Evolution of Robust Solutions

Nicola Milano\*, Jônata Tyska Carvalho\*#, Stefano Nolfi\*

\*Institute of Cognitive Sciences and Technologies
National Research Council (CNR),
Roma, Italia,

#Center for Computational Sciences (C3),
Federal University of Rio Grande (FURG),
Av. Italia, km 8, Rio Grande, Brasil
jonata.carvalho@istc.cnr.it

**Abstract**

Previous evolutionary studies demonstrated how evaluating evolving agents in variable environmental conditions enable them to develop solutions that are robust to environmental variation. We demonstrate how the robustness of the agents can be further improved by exposing them also to environmental variations throughout generations. These two types of environmental variations play partially distinct roles as demonstrated by the fact that agents evolved in environments that do not vary throughout generations display lower performance than agents evolved in varying environments independently from the amount of environmental variation experienced during evaluation. Moreover, our results demonstrate that performance increases when the amount of variations introduced during agents' evaluation and the rate at which the environment varies throughout generations are moderate. This is explained by the fact that the probability to retain genetic variations, including non-neutral variations that alter the behavior of the agents, increases when the environment varies throughout generations but also when new environmental conditions persist over time long enough to enable genetic accommodation.

*Keywords—environmental variations; evolvability; stability; artificial evolution.*

# Introduction

The last two decades have seen an increasing recognition of the role of environmental variations in evolution.

The interaction between environmental conditions and the expression of genetic variation influences the evolutionary dynamics. Genes influencing a trait in one environment may not be important in a different one (Viera et al., 2000). Mutations often have environment-dependent effects (Kawecki, 1994; Szafraniec, Borts and Korona, 2001). The environmental conditions influence the genetic interactions among traits, i.e., the correlation between the genetic influences on a trait and the genetic influences of another trait, which are known to influence the evolutionary dynamics (Sgro and Hoffmann, 2004). For instance, the genetic correlations among certain traits can be positive in an environment and negative in another one. Consequently environmental variations influence evolutionary trajectories in populations (Sgro and Hoffmann, 2004).

Moreover, as stressed by West-Eberhard (2003), phenotypic variation arises not only as a result of genetic variations but also as a result of environmental variations. "Environmentally induced phenotypic changes can give rise to adaptive evolution as readily as mutational induced changes; both are equally subject to genetic accommodation." (West-Eberhard, 2003, pp.498).

In this paper we analyze the impact of environmental variation on the evolution of neuro-controlled agents situated in an external environment. More specifically we analyze whether agents evolved in environmental conditions that vary over generations outperform agents evolved in non-varying environments. The obtained results demonstrated that indeed agents evolved in varying environments outperform agents evolved in environments that do not vary throughout generations independently from the amount of environmental variation experienced by evolving individuals.

This study is related to the area of optimization in dynamic environment (Jin and Branke, 2005; Cruz, Gonzalez and Pelta, 2011) that focus on how evolving solutions can cope with optimization problems that are dynamic and change over time stochastically and/or periodically (see also Kashtan, Noor and Alon, 2007; O′Donnell at al., 2014; Janssen et al., 2016). However, the objective of these studies is to evolve flexible solutions, i.e. agents capable of adapting to new environmental conditions during a certain number of generations, rather than robust solutions, i.e. agents capable of operating effectively in new environmental conditions immediately without the need to further evolve. Consequently, also methodological issues differ. For example, the inclusion of mechanisms that preserve population diversity is clearly important for the evolution of flexible agents capable of adapting to the new environmental conditions in few generations but is not necessarily important for the evolution of robust agents. Moreover, the formalization of a method for measuring the speed with which agents adapt to the new environmental conditions is crucial for studying the evolution of flexible agents but is not relevant for the study of robust agents.

Previous research demonstrated how exposing evolving candidate solutions to (deliberately introduced) variations and by averaging the fitness obtained in varied conditions can successfully lead to robust solutions in different domains: electronic circuits robust to temperature variations (Thompson and Layzell, 2002), fault tolerant neural networks (Sebald and Fogel, 1992), job shop scheduling (Tjornfelt-Jensen and Hansen, 1999), flight control under changing conditions (Blythe, 1998), robot control in varying environmental conditions (Nolfi et al. 1994; Jacoby, 1997). In this work we investigate the relationship between the amount of environmental variation experienced by each evolving candidate solutions, the amount of environmental variation eventually occurring throughout generations, and the capability of the evolved agents to operate effectively in varying environmental conditions. The obtained results indicate that the best performance is obtained when candidate solutions are subjected to environmental variations both while they are evaluated and throughout generations. Moreover, our results indicate that the performance is greater when amount of variation experienced by each candidate solutions is moderate (i.e. when candidate solutions are evaluated for a limited number of trials in which the environmental conditions vary) and when the environment change throughout generations at a moderate rate (i.e. when the environment does not vary every generation but only every N generations).

The rates at which evolving candidate solutions vary throughout generations at the genetic and behavioral levels are also greater in the condition in which the environment vary at a moderate rate throughout generations than in the conditions in which the environment remain fixed or varies every generation. Therefore the advantage provided by the condition in which the environment varies at a moderate rate throughout generations can be explained by considering that this condition combines the advantage of experiencing varied environmental conditions throughout generations with the need to experience constant environmental conditions long enough to enable the generation and the selection of genetic variations that are adaptive.

**Method**

Agents are evolved for the ability to solve an extended version of the double-pole balancing problem introduced by Wieland (1991) that became a commonly recognized benchmark for nonlinear control (Igel, 2003; Khan et al., 2010; Wiestra et al., 2014). As in the standard version of the problem, agents are constituted by carts with two poles of different length attached on their top through passive joints (Figure 1, left). In our extended version of the problem, the plane over which the agents are situated can be inclined of variable angles and is constituted by materials that generate variable friction forces between the plane and the cart. In the standard problem, instead, the inclination of the plane and the cart/plane friction are always zero.

The characteristics of the environment that vary are the inclination and the friction of the plane and the initial state of the cart. Consequently, robust agents should display an ability to balance the poles irrespectively from the characteristics of the plane and the initial state of the cart.

Agents are provided with a three layers neural network with five sensory neurons, ten internal neurons with recurrent connections, and one motor neuron (Figure 1, right). The sensory neurons encode the position of the cart ($x$), the angular position of the two poles ($\theta_1$ and $\theta_2$), the inclination of the plane ($\alpha$), and the friction coefficient of the plane ($\mu_c$). The activation state of the motor neuron is normalized in the range [-10.0, 10.0] N and is used to set the force (F) applied to the cart along the $x$ axis.

The connection weights of the neural network, that determine the behavior of the agents, are encoded in artificial genotypes and evolved through a steady state evolutionary algorithm, a method widely used to evolve embodied agents (Nolfi et al., 2016)

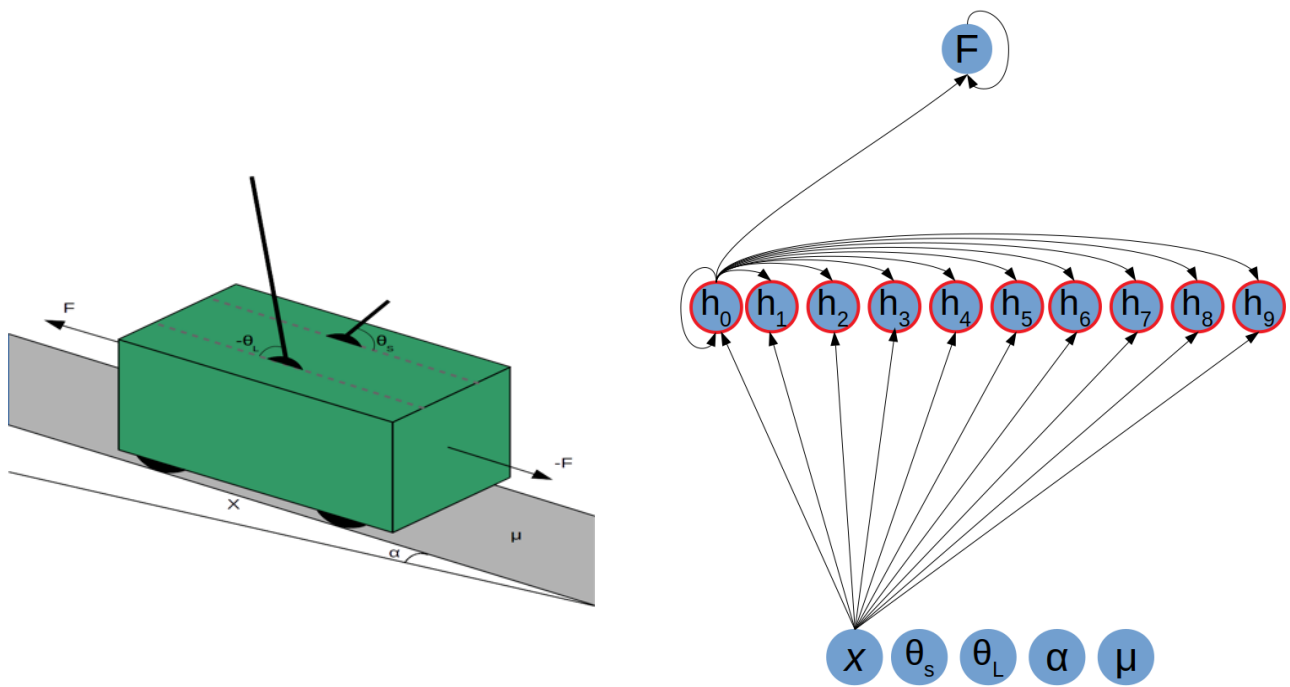

Figure 1. Left: Schematization of the extended double-pole balancing problem. See text for explanation. Right: The architecture of the neural network controller. The circles shown in the bottom, central and top part of the figure represents the sensory, internal and motor neurons, respectively. Red circles represent the biases. The arrows represent connections. For sake of clarity, only the connection departing from the first sensory and the first internal neurons are displayed.

Each agent is evaluated for NT trials that vary with respect to the characteristics of the plane and with respect to the initial state of the cart and the poles. More specifically, at the beginning of each trial the inclination of the plane ($\alpha$), the friction coefficient between cart and plane ($\mu_c$), the initial position of the cart on the plane ($x$), the velocity of the cart ($\dot{x}$), the angular position of the two poles ($\theta_1$ and $\theta_2$) and the angular velocity of the two poles ($\dot{\theta}_1 \wedge \dot{\theta}_2$) are set to values selected randomly within the following ranges: [0.0, 0.2617], [0.0, 0.30], [-1.5, 1.5], [-1.2, 1.2], [-0.1047, 0.1047], [-0.1350, 0.1350], respectively.

Trials terminate after 1000 steps or when the angular position of one of the two poles exceeded the range [-36º, 36º] or the position of the cart exceed the range [-2.4, 2.4] m.

The fitness of the agent during a trial ($f_i$) corresponds to the fraction of time steps in which the agent maintains the cart and the poles within the allowed position and orientation ranges. The total fitness is obtained by averaging the fitness obtained in the NT trials.

The performance of an agent, i.e. the ability of an agent to solve the problem in variable environmental conditions, is measured by post-evaluating evolved agents for 1000 trials in which the characteristics of the environment and the initial state of the cart are set randomly with a uniform distribution in the ranges described above. Although the number of different environmental conditions that an agent can encounter is practically infinite, measuring the performance of the agents on 1000 randomly different environmental conditions provides a good estimate of the robustness of the agents, i.e. of the ability of the agents to solve the problem in all possible environmental conditions.

The experiments have been replicated in a fixed environmental condition in which the environmental conditions do not vary throughout generations, in a varying environmental condition in which the environmental conditions vary every generation, and in an intermediate condition in which the environmental conditions vary every 100 generations. The environmental conditions of each trial are stored in a vector of 8 values that encode the characteristics of the plane and the initial states of the cart and of the poles. In the fixed experimental condition, the *NT*x8 matrix used during the *NT* corresponding trials is chosen randomly once and is maintained constant during the entire evolutionary process. In the always-varying experimental conditions, the *NT*x8 matrix is generated randomly every generation. In the intermediate conditions, the matrix is re-generated randomly every N generations.

The amount of environmental variability experienced by each candidate solution can be manipulated by varying the number of trials (NT). The amount of environmental variability experienced throughout generations can be manipulated by varying the number of generations after which the characteristics of the environment are varied.

The evolutionary process is continued until a maximum number of agents/environmental evaluations is performed. This enable to compare the results of different evolutionary experiments by maintaining the computational cost approximately constant given that the computational cost of the other operations is negligible with respect to the cost of evaluating agents' behavior. For each experiment we ran 30 replications.

A detailed description of the method is provided in the Appendix (Section 1-4). The experiments performed can be replicated by downloading and installing FARSA simulator (Massera et al., 2014) from "https://sourceforge.net/projects/farsa/" and the experimental plugin from http://laral.istc.cnr.it/res/moderateenvvar/plugin.zip.

**Moderate environmental variation promotes the evolution of better solutions**

As expected, evolved agents evaluated for multiple trials in which the environmental conditions vary outperform agents evolved in non-variable environmental conditions, i.e. agents evaluated for a single trial, see the appendix.

The optimal number of trials, i.e. the number of trials that enables the evolving agents to achieve the best performance during the post-evaluation test made on 1000 randomly different environmental conditions, varies depending on whether the environment varies also throughout generations or not. Indeed, the best performance is obtained with 200 trials in the experiments in which the environment does not vary over generations and with 25 trials in the experiments in which the environment varies every generation or every 100 generations. The fact that the optimal number of trials is greater in the case of the experiments in which the environment does not vary over generations indicates that the lack of environmental variation throughout generations can be compensated in part by increasing the variations of the conditions in which agents are evaluated (i.e. the number of trials). The fact that the lack of environmental variation throughout generations can be compensated only partially by increasing the variation of the conditions in which agents are evaluated is demonstrated by the fact that the utilization of more than 200 trials in the fixed environmental condition produces a reduction of the obtained performance (see table 1 in appendix) and by the fact that agents evolved in environments that vary every generation or every 100 generations outperform the agents evolved in environments that do not vary throughout generations independently from the number of trials (Welch's t-test p-value $< 10^{-3}$, comparison performed by using the parameters that resulted optimal in each condition).

The performance achieved in the intermediate condition in which the environment varies every 100 generations is significantly better than the performance achieved in the fixed and always varying experimental conditions (Welch's t-test with Bonferroni correction p-value $< 0.01667$, comparison performed by using the parameters that resulted optimal in each condition). The condition that leads to the best performance, therefore, is that in which the agents are evaluated for a limited number of trials in different environmental conditions and in which the environment varies throughout generations at a moderate rate (i.e. every 100 generations and not every generation). Indeed, the intermediate condition in which the environment varies every 100 generations outperforms the fixed and always-varying experimental conditions both the experiments continued for 50 millions evaluations and in the experiments continued for 5,000 generations (Welch's t-test with Bonferroni correction p-value $< 0.0123$). Moreover, the intermediate condition in which the environment varies every 100 generations outperforms the fixed and always-varying experimental conditions independently from whether the number of trials is set to 25, 50, or 200 (Section 5 of the Appendix, Welch's t-test with Bonferroni correction p-value $< 0.0098$, comparison performed by using the parameters that resulted optimal in each condition).

The fact that the performance of the agents is greater in the experiments in which the environment changes at a moderate rate over generations is confirmed by the analysis of the performance achieved in a series of experiments in which we systematically varied the rate at which the environment changes over generations (see Figure 2). This data has been obtained by keeping the other parameters equal. The mutation rate was set to 1% (the value that resulted optimal in all conditions, see Section 5 of the Appendix). The number of trials and the stochasticity level (a parameter controlling the strength of elitism, see section 3 of the Appendix) were set to 50 and 0%, respectively, i.e. to values that produced high performance in all conditions (see Section 5 of the Appendix). The performances obtained when the environment varies every 10, 25, 50, 100 and 200 generations do not differ among themselves (Kruskal-Wallis ANOVA p-value=0.09) but they differ statistically from the other experimental conditions (Kruskal-Wallis ANOVA p-value p-value $< 0.001$, for all comparisons). In the rest of the paper, we will use the term intermediate experimental condition to refer to the experiments in which the environment varies every 100 generations since it corresponds approximately to the average value of the best conditions.

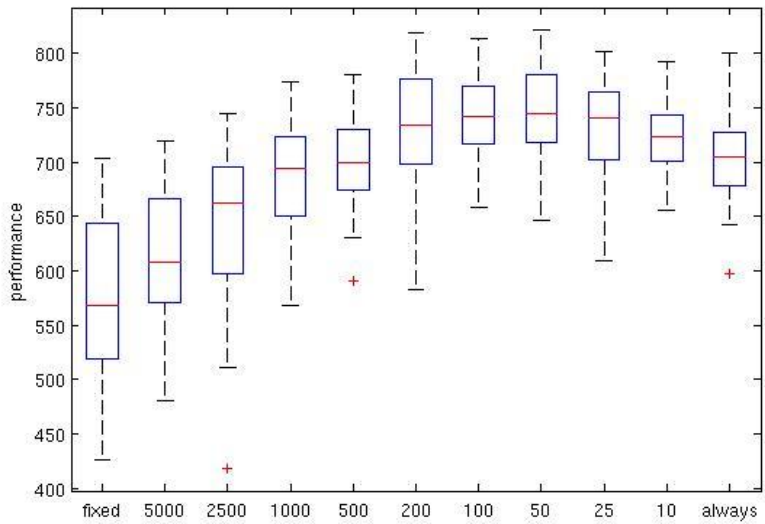

Figure 2. Performance in the fixed experimental condition, in eight intermediate conditions in which the environment varies every 5000, 2500, 1000, 500, 200, 100, 50, 25, and 10 generations (5000-10), and in the always varying experimental condition. Boxes represent the inter-quartile range of the data and horizontal lines inside the boxes mark the median values. The whiskers extend to the most extreme data points within 1.5 times the inter-quartile range from the box. Circles indicate the outliers. The evolutionary process was continued for 50 millions evaluations. Average results of 10 replications.

The beneficial effect of a moderate rate of environmental variations throughout generation persists on the long term as demonstrated by the fact that the agents evolved in the intermediate condition outperform agents evolved in fixed and always varying conditions both after 50 and 100 million evaluations (Fig. 3, Welch's t-test with Bonferroni correction, p-value < 0.0095). Part of these data were previously reported in Milano, Carvalho and Nolfi (2017).

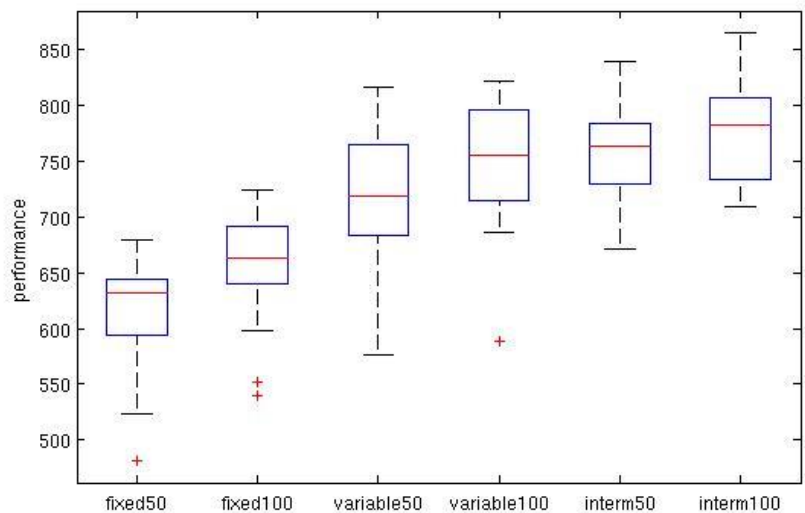

Figure 3 Performance of agents evolved in the fixed environmental condition (fixed), in the condition in which the environment change every generations (variable), and in the condition in which the environment change every 100 generations (interm) after 50 and 100 millions evaluations. Results obtained with the best parameters, i.e. mutation rate 1% in all conditions, number of trials 200 in the fixed environmental condition and 25 in the intermediate and always variable experimental conditions, stochasticity 30%, 0%, and 20% in the case of the fixed, intermediate and always variable experimental conditions. Boxes represent the inter-quartile range of the data and horizontal lines inside the boxes mark the median values. The whiskers extend to the most extreme data points within 1.5 times the inter-quartile range from the box. Circles indicate the outliers.

**On the role of fortunate specific environmental conditions**

A possible hypothesis that could explain why agents evolved in varying environmental conditions achieve better performance is that environmental variations enable evolving agents to experience, sooner or later, specific environmental conditions that promote the evolution of better solutions.

This hypothesis is based on evidences collected in incremental evolutionary experiments in which the problem and/or the environment become progressively more challenging throughout generations (Mikkulainen and Gomez, 1997; Nolfi and Floreano, 1999). The possibility to experience easier conditions during the first generations facilitates the evolution of solutions that can then be adapted to solve also more challenging problems. Moreover the hypothesis is based on evidences collected in neural network learning literature that indicate that the relative distribution of qualitatively different stimuli in the training set strongly affect the outcome of the learning process (Hare and Elman, 1995; Zhou and Liu, 2006). Environmental variations in our experiments are stochastic and consequently cannot lead to a progressive complexification of the adaptive problem. On the other hand, one could hypothesize that that the continuous variation of the environmental conditions can generate, sooner or later, sufficiently easy environmental conditions or fortunate environmental conditions that can boost evolution.

To verify this hypothesis we measured how the performance of the same agents evolved in randomly different environmental conditions vary over generations. The analysis was conducted on 30 populations of agents evolved in the intermediate experimental condition in which the environment varied every 100 generations after one, five, and then ten thousand generations. One hundred copies of these populations were evolved for 500 generations in 100 different corresponding environments for 50 trials. The performances of these populations were post-evaluated every 100 generations on 729 trials in which agents were exposed to systematically varied environmental conditions (see Section 6 of the Appendix).

As can be seen, the performance of the agents after one and five thousand generations increases of a similar amount in all environmental conditions (Figure 4 top and middle pictures). Later on, i.e. after ten thousands generations, performance variations become much smaller (Figure 4 bottom, notice that the scale used for displaying variation is one order of magnitude smaller in the case of the bottom picture). Moreover, the ability of the agents to solve the 729 trials increases or decreases slightly while they are evolved in different environments, although the number of environments that lead to performance increase is higher than the number of environments that lead to performance loss.

This data does not show evidences of fortunate specific environmental conditions capable of boosting the evolutionary process. During the initial evolutionary phase, the evolving agents improve their ability of a similar amount in all environmental conditions. Moreover, the evolving agents improve their ability in the majority of the environmental conditions also during successive evolutionary phases.

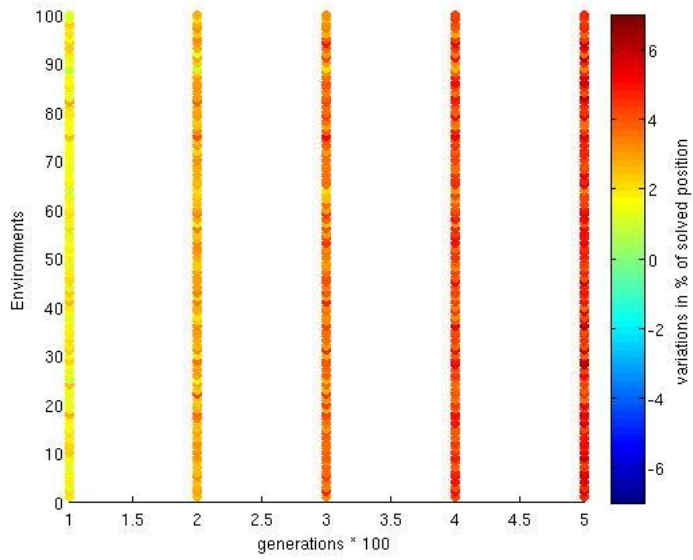

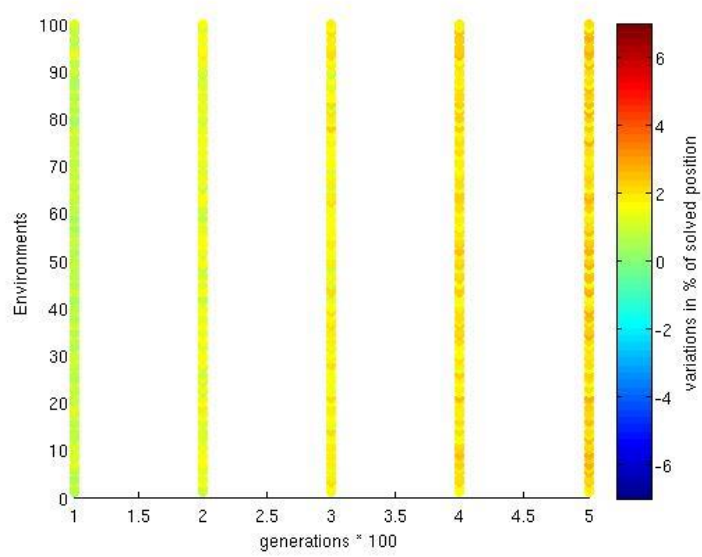

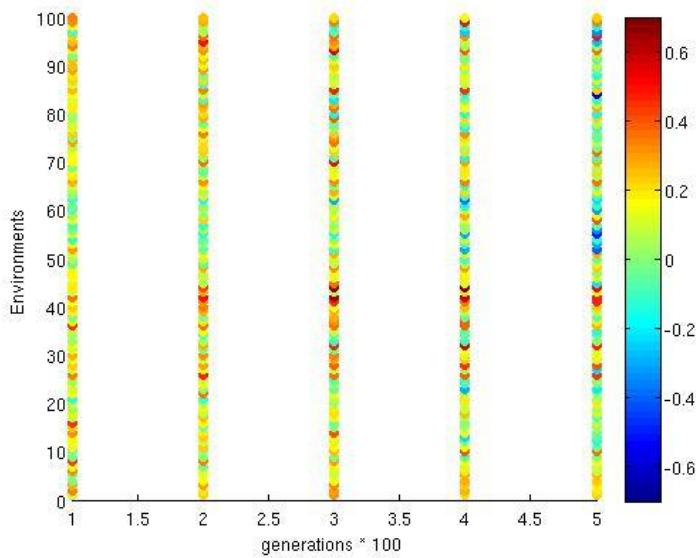

Figure 4. Fraction of additional trials solved every 100 generations by populations of agents evolved for 500 generations in 100

different environments. The top, middle, and bottom pictures show the results obtained by analyzing the agents evolved in the intermediate condition after one, five, and then ten thousand generations, respectively. Data averaged over 10 replications of the experiment. Columns correspond to data after 100, 200, 300, 400, and 500 generations. Lines indicate performance variations observed in each of the 100 different environments.

**Environmental variation increases the rate at which evolving agents change throughout generations**

Another factor that could explain why moderate environmental variation throughout generations promotes the evolution of better agents is constituted by the possible impact that environmental variation has on the rate at which evolving agents change throughout generations.

To understand the possible role of this factor we should consider that adaption depends on the generation of phenotypic changes and on the retention of the changes that are adaptive. As stressed by West-Eberhard (2003), phenotypic variations arise not only as a result of genetic variations but also as a result of environmental variations. Moreover, adaptations can arise both as a result of mutational induced changes and as a result of environmentally induced changes since both are subjected to genetic accommodation (West-Eberhard, 2003). The occurrence of environmentally induced changes resulting from environmental variations in addition to genetically induced changes induced by mutations can therefore enable evolving agents to vary more throughout generations which in turn can facilitate the discovery of better solutions.

The fact that genetic accommodation requires time can potentially explain why the advantage of environmental variations is greater at intermediate condition in which the environment varies every 100 generations. Too frequent environmental variations (e.g. environments varying every generation) prevent the stability that is necessary to enable genetic accommodation. On the other hand, environmental variations that occur too rarely provide less opportunity for change. An intermediate variation rate can thus represent an optimal tradeoff between the contrastive needs of variation and stability. This hypothesis is in line with West-Eberhard's claim that the most important reason that explains why environmentally induced changes are evolutionarily important is indeed their time persistence:

*Perhaps the most compelling argument for the superiority of environmental induction over mutations in term of recurrence and persistence has to do with the inexorable persistence of an environment immune to natural selection: environmental inducers might be not only immediately widespread without necessity for positive effects on fitness sufficient to spread them due to differential reproduction of their bearers (selection), but they are inexorably present.*

West-Eberhard (2003), pp. 504

A possible hypothesis that could explain why agents evolved in environments that vary over generations outperform agents evolved in non-varying environment and why agents evolved in environments that vary every 100 generations outperform agents evolved in environments that vary every generation is that the rate at which the evolving agents change throughout generation is greater when the environment varies at a moderate rate (i.e. every 100 generations).

To verify the impact of the rate at which the environment varies through generations on the rate at which the agents vary throughout generations at the behavioral level we compared the best evolved agent and its ancestor of 100 generations before every 100 generations. The comparison was made by post-evaluating the agents for 729 trials during which they were exposed to systematically varied environmental conditions (see Section 6 of the Appendix) and by counting the number of trials in which the agent of generation N+100 manage to balance the poles while its ancestor of generation N fails or vice versa. To ensure a proper comparison, the analysis was performed on experiments terminated after 5000 generations. As in the case of the analysis reported in Figure 2, the number of trials was set to 50, the mutation rate to 1%, and the stochasticity level to 0%.

As expected, the rate of variation decreases throughout generation as a result of the evolution of better and better agents (Figure 5). Interestingly, the agents evolved in the intermediate experimental condition in which the environment varies every 100 generations (blue line) vary more at the behavioral level throughout generations than the agents evolved in the fixed experimental condition (black line) and the always-varying experimental condition (red line) (Kruskal Wallis test p value $< 3 \cdot 10^{-6}$).

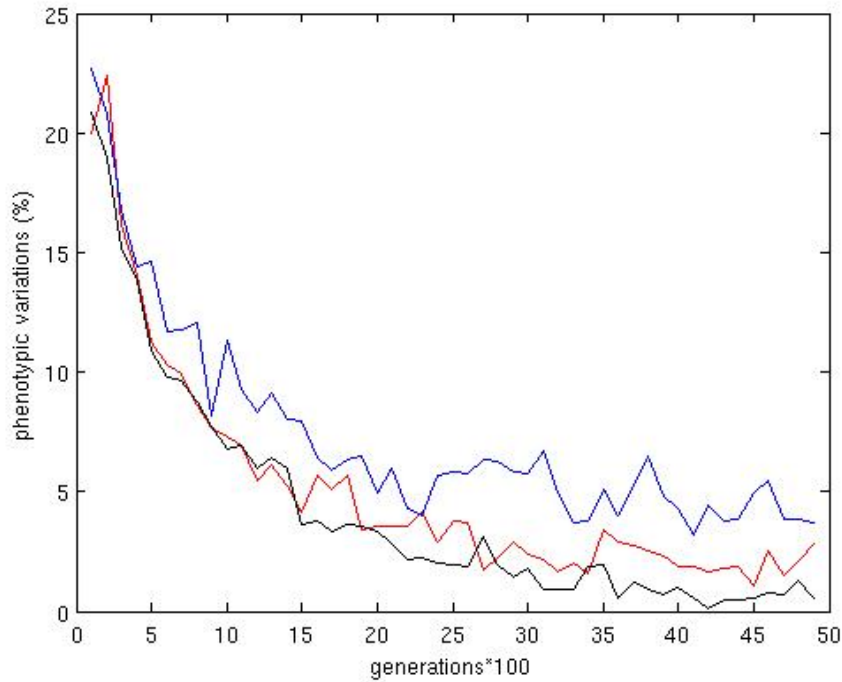

Figure 5. Fraction of trials in which the ability to solve the double-pole navigation problem varies every 100 generations. The black, blue, and red curves display the data obtained in the experiments carried out in the fixed, intermediate (in which the environment changes every 100 generations), and always-varying experimental conditions, respectively. Data obtained by analyzing the evolutionary lineage of the fittest evolved individual of each experiment. Each curve displays the average results of 30 replications terminated after 5000 generations

The analysis of how the evolving agents vary throughout generations at the genetic level indicates that, also in this case, the agents evolved in the intermediate condition accumulate more variations throughout generations than the agents evolved in the fixed and always varying experimental conditions (Figure 6, Kruskal Wallis test p value $< 0.005$). Data indicate the fraction of connection weights and biases that differ between the best evolved agent and its ancestor of 500 generation before every 500 generations. Also in this case, to ensure a proper comparison, the analysis was performed on experiments continued for 5000 generations.

The rate at which agents change throughout generations can also be influenced by the mutation rate which however is always 1% in the case of the experiments reported in this section. Moreover, the results of the experiments carried out by varying systematically the mutation rate and the other parameters (see the Appendix) show that the agents evolved in the intermediate condition, in which the environment vary every 100 generations, outperform the agents evolved in the fixed and always varying experimental conditions independently from the combination of parameters. Consequently, the improvement in performance that can be gained by exposing evolving agents to environments that vary every 100 generations cannot be gained by adjusting the other parameters.

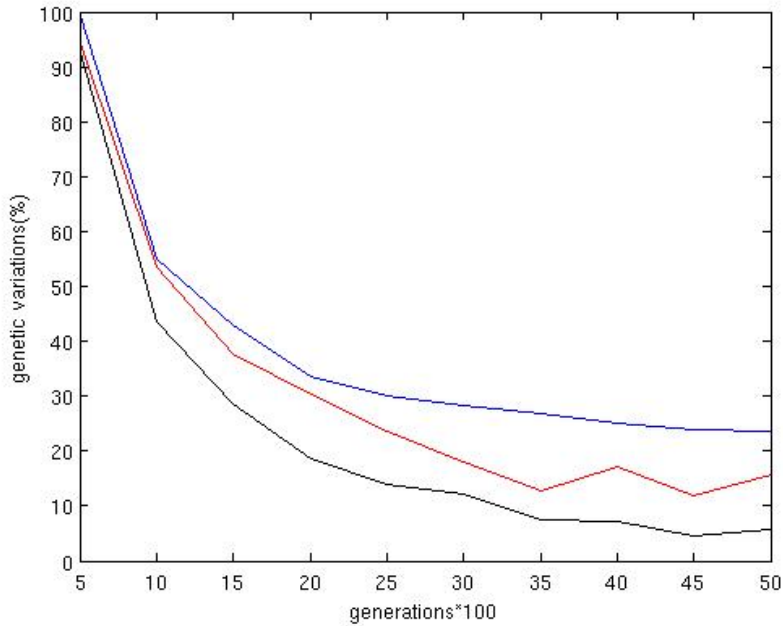

Figure 6. Fraction of genes varied every 500 generations. The black, blue, and red curves display the data obtained in the experiments carried out in the fixed, intermediate (in which the environment changes every 100 generations), and always-varying experimental conditions, respectively. Data obtained by analyzing the evolutionary lineage of the fittest evolved individual of each experiment. Each curve displays the average results of 10 replications of each experiment.

The analysis of these data indicates a significant correlation between the performance of evolving agents and the rate with which agents vary throughout generations at the behavioral (Spearman r= 0.52 $p<10^{-7}$) and genetic level (Spearman r= 0.53 $p<10^{-8}$).

These results confirm the hypothesis that environmental variations increase the rate at which agents change throughout generations at the genetic and behavioral level and the hypothesis that the rate at which the agents change is greater when the environment varies at moderate rate throughout generation than when the environment vary at a high rate throughout generations.

The presence of a significant correlation between the performance of the evolved agents and the rate at which agents vary throughout generations, at the genetic and behavioral level, indicates the improvement in performance obtained by evolving agents in environment that vary every 100 generations is due, at least in part, to the fact that moderate environmental variations promote phenotypic variations, i.e. facilitate the retention of genetic variations that alter the behavior of the agents.

**Conclusions**

Previous research reviewed in the introduction demonstrated how evaluating candidate solution in variable environmental condition enables to evolve solutions displaying a certain level of robustness with respect to environmental variation. Our results demonstrate that the introduction of environmental variations also throughout generations promote the evolution of better solutions.

The analysis of the results obtained by systematically varying the amount of environmental variations introduced during agents'evaluation and throughout generations demonstrates that these two type of environmental variations play partially distinct roles as demonstrated by the fact that agents evolved in environments that do not vary throughout generations display lower performance than agents evolved in environments that vary throughout generations independently from the amount of environmental variation experienced by each evolving agent.

Moreover, our results demonstrate that best performance is obtained when the amount of variations introduced during agents' evaluation and the rate at which the environment varies throughout generations are moderate. By analyzing the rate at which evolving agents change throughout generations in the different experimental conditions we observed that the rate at which agents vary at the genetic and behavioral level throughout generations is greater in the experiments in which the environment varies throughout generations, especially when the rate of environment variation over generation is moderate. Overall this indicates that the beneficial effect of varying the environment at a moderate rate is due to the fact that it maximizes the rate at which evolving agents change evolutionarily.

From an evolutionary computation perspective, our results indicate that introducing moderate environmental variations both during the evaluation of candidate solutions and throughout generations permits to maximize performance. Agents' evaluation represents the major computational cost in evolutionary computation. Consequently, the fact that the optimal number of different environmental conditions in which candidate solution should be evaluated is small, especially when the environment varies also throughout generations, has important practical implications. In the case of the experiments reported in this paper this implies that the number of trials that should be used to evaluate each candidate solution can be small. This means that the computational cost of evolving agents in widely varying environmental conditions can be manageable despite the cost of evaluating each candidate solution in all possible environmental conditions is unsustainable.

From an evolutionary biological perspective, our results provide additional support to the hypothesis of West-Eberhard (2003) that environmentally induced changes play an important role in evolution thanks in particular to their tendency to persist over time independently from their adaptive nature.

**Appendix**

**1. The agent and the environment**

The cart has a mass of 1Kg. The long pole and the short pole have a mass of 1.0 and 0.1 Kg and a length of 0.5 and 0.05 m, respectively. The cart can move along one dimension within a track of 4.8 m. It is provided with five sensors that encode the current position of the cart on the track ($x$), the current angle of the two poles ($\theta_L$ and $\theta_S$) with respect to the cart, the angle of the inclined plane (α) and the friction coefficient (μ). The motor controls the force (F) applied to the cart along the x axis. The goal of the agent is to move so to maintain the angle of the poles and the position of the cart within a viable range (see below).

The behavior of the agent has been simulated on the basis of equations 1-5. This is an extended version of the equations proposed by Florain (2007) for the standard problem, in which the inclination of the plane and the friction between the cart and the plane were not considered.

$$\ddot{x} = \frac{F + \mu_c \widetilde{M} g + M_c g \sin \alpha + \sum_{i=1}^{n} \widehat{F}_i}{M_c + \sum_{i=1}^{n} \widetilde{m}_i} \tag{1}$$

$$\ddot{\theta}_i = -\frac{3}{4 l_i} \left( \ddot{x} \cos \theta_i - g \sin \theta_i + \frac{\mu_p \dot{\theta}_i}{m_i l_i} \right) \tag{2}$$

$$\widehat{F}_i = \mu_c \left[ \frac{3}{4} m_i g \sin^2 \theta_i - \frac{3\mu_p}{4 l_i} \dot{\theta}_i \sin \theta_i + m_i l_i \dot{\theta}_i^2 \cos \theta_i \right] - \frac{3}{4} \left[ m_i g \sin \theta_i \cos \theta_i + \mu_p \dot{\theta}_i \cos \theta_i \right] + m_i l_i \dot{\theta}_i^2 \sin \theta_i \tag{3}$$

$$\tilde{m}_i = \frac{3}{4}[cos^2\theta_i - \mu_c cos\theta_i sin\theta_i] \qquad (4)$$

$$\tilde{M} = M_c cos\alpha + \sum_{i=1}^{n} m_i \qquad (5)$$

where $n$ is the number of poles on the cart, $g$ is the acceleration due to gravity, $m_i$ and $l_i$ are the mass and the half length of the $i^{th}$ pole, $M_c$ is the mass of the cart, $\mu_c$ is the coefficient of friction of the cart on the track, $\mu_p$ is the coefficient of friction for the $i^{th}$ hinge, $F$ is the force applied to the cart, $\hat{F}_i$ is the effective force from the $i^{th}$ pole on the cart, $\tilde{m}_i$ is the effective mass of the $i^{th}$ pole, and $\tilde{M}$ is the effective mass of the cart.

## 2. The neural network controller of the agent

The controller of the agent is constituted by a neural network with five sensory neurons, ten internal neurons with recurrent connections, and one motor neuron. The sensory neurons are fully connected with the internal neurons, and the internal neurons are fully connected with the motor neurons and the internal neurons.

The sensory neurons encode the position of the cart ($x$) expressed in meters, the angular position of the two poles ($\theta_1$ and $\theta_2$) radians, the inclination of the plane ($\alpha$) radians, and the friction coefficient of the plane/cart ($\mu_c$). The state of all sensors is normalized in the range [-0.5, 0.5]. The activation state of the motor neuron is normalized in the range [-10.0, 10.0] N and is used to set the force applied to the cart. The state of the sensors, the activation of the neural network, the force applied to the cart, and the position and velocity of the cart and of the poles are updated every 0.01 s.

The neural network's architecture is fixed. The activation state of the internal and motor neurons is updated on the basis of the logistic function. The connection weights and the biases of the network are encoded in agent's genome and evolved. More specifically each genome consists of a vector of 171 x 8 = 1368 bits that encode the 160 connection weights and the 11 biases of the corresponding neural network controller.

## 3. The evolutionary method

The evolutionary algorithm consists of a simple ($\mu + 1$) evolutionary strategy Rechenberg (1973) that operates on the basis of populations formed by $\mu$ parents. During each generation, each parent generates one offspring, the parent and the offspring are evaluated, and the best $\mu$ individuals are selected as new parents. When the environmental conditions do not change with respect to the previous generation, the fitness of the parent is set equal to the fitness measured during previous evaluations and the evaluation of the parents is skipped.

The genome of the initial population is composed by a matrix of ($\mu$ x 1368) bits that are initialized randomly. Each block of 8 bits is converted into a floating-point number in the range [-5.0, 5.0] that is used to set the weight of the corresponding connection or bias of the neural network controller. Offspring are generated by creating a copy of the genotype of the parent and by subjecting each bit to mutation with a *MutRate* probability. Mutations are realized by flipping the mutated bit.

The selection pressure is regulated by adding to the fitness of individuals a value randomly selected in the range [-Noise, Noise] with a uniform distribution (Jin and Branke, 2005), where Noise corresponds to the theoretical maximum fitness multiplied by the value of the Stochasticity parameter. When Stochasticity is set to 0.0 only the best µ individuals are allowed to reproduce. The higher the level of stochasticity, the higher the probability that the worse individuals reproduce is.

This method requires to set two parameters: MutRate and Stochasticity. To identify the optimal values of the parameters we carried a series of control experiments in which the two parameters were varied systematically (see the following sections). The method operates well on small populations, e.g. populations formed by 100 individuals (Whitley, 2001).

## 4. Fitness function and performance measure

Agents are evolved for the ability to solve an extended version the non-markovian version of the double-pole balancing problem Wieland (1991) in which the plane over which the agent is situated can be inclined of varying angles and the friction between the plane and the cart varies.

Each evolving agent is evaluated for NT trials that vary with respect to the characteristics of the plane and with respect to the initial state of the cart and the poles. More specifically, at the beginning of each trial the inclination of the plane ($\alpha$), the friction coefficient between cart and plane ($\mu_c$), the initial position of the cart on the plane ($x$), the velocity of the cart ($\acute{x}$, the angular position of the two poles ($\theta_1$ and $\theta_2$) and the angular velocity of the two poles ($\acute{\theta}_1 \wedge \acute{\theta}_2$ are set to values selected within the following ranges, respectively: [0.0, 0.2617], [0.0, 0.30], [-1.5, 1.5], [-1.2, 1.2], [-0.1047, 0.1047], [-0.1350, 0.1350].

Trials terminate after 1000 steps or when the angular position of one of the two poles exceeded the range [-36º, 36º] or the position of the cart exceed the range [-2.4, 2.4] m.

The fitness of the agent during a trial ($f_i$) corresponds to the fraction of time steps in which the agent maintains the cart and the poles within the allowed position and orientation ranges and is calculated on the basis of the following equation:

$$f_i = \frac{t}{1000} \tag{6}$$

where t is the time step in which the cart or the pole angles exceeded their allowed range or 1000 in case they are maintained in the range until the end of the trial. The total fitness (F) is calculated by averaging the fitness obtained during the different trials:

$$F = \frac{\sum_{i=1}^{NT} f_i}{NT} \tag{7}$$

where *NT* indicates the number of trials.

The performance of an agent, i.e. the ability of an agent to solve the problem in all environmental conditions, is measured by post-evaluating agents for 1000 trials in which the characteristics of the environment and the initial state of the cart are set randomly with a uniform distribution in the ranges described above. To ensure a proper comparison the same matrix of 1000 x 8 values generated randomly is used for post-evaluating agents evolved in different experiments. Performance measures thus rate how the evolving agents are capable of generalizing their abilities to solve the double-pole problem in environmental conditions that differ from the conditions experienced during evolution.

## 5. Performance achieved with systematically varied parameters

Table 1, 2 and 3 reports the results obtained by systematically varying the number of trials, the mutation rate and the stochasticity level in experiments carried out in the fixed, always-varying and intermediate experimental condition in which the environment does not vary throughout generations, varies every generation, and vary every 100 generations, respectively. The population size is always set to 100. The evolutionary process was continued for 50 millions evaluations($n_{individuals} \times n_{generations} \times n_{trials}$. Each number indicates the average results of 10 replications.

| 1 Trial | Stoch 0% | Stoch 10% | Stoch 20% | Stoch 30% | Stoch 40% |
|---|---|---|---|---|---|
| Mut 1% | 78.9 | 85.6 | 95.3 | 86.3 | 75.3 |
| Mut 2% | 68.3 | 75.3 | 81.3 | 74.2 | 70.2 |
| Mut 4% | 55.3 | 63.2 | 53.3 | 56.8 | 55.2 |
| 50 Trials | Stoch 0% | Stoch 10% | Stoch 20% | Stoch 30% | Stoch 40% |
| Mut 1% | 562.9 | 578.6 | 629.8 | 613.4 | 646.4 |
| Mut 2% | 606.9 | 628.4 | 625.7 | 645.2 | 619.6 |
| Mut 4% | 537.2 | 504.0 | 537.3 | 506.8 | 525.2 |
| 100 Trials | Stoch 0% | Stoch 10% | Stoch 20% | Stoch 30% | Stoch 40% |
| Mut 1% | 625.6 | 595.9 | 622.0 | 615.9 | 634.1 |
| Mut 2% | 658.1 | 629.9 | 623.3 | 574.5 | 606.5 |
| Mut 4% | 479.2 | 442.6 | 506.1 | 477.0 | 463.0 |
| 150 Trials | Stoch 0% | Stoch 10% | Stoch 20% | Stoch 30% | Stoch 40% |
| Mut 1% | 649.9 | 639.7 | 653.9 | 644.1 | 643.9 |
| Mut 2% | 601.6 | 605.3 | 624.2 | 629.8 | 602.9 |
| Mut 4% | 431.8 | 461.8 | 466.1 | 462.0 | 427.0 |
| 200 Trials | Stoch 0% | Stoch 10% | Stoch 20% | Stoch 30% | Stoch 40% |
| Mut 1% | 641.6 | 644.2 | 634.0 | **658.7** | 636.8 |
| Mut 2% | 584.5 | 592.7 | 594.2 | 572.1 | 584.8 |
| Mut 4% | 424.5 | 442.7 | 444.2 | 432.1 | 427.8 |
| 300 Trials | Stoch 0% | Stoch 10% | Stoch 20% | Stoch 30% | Stoch 40% |
| Mut 1% | 612.8 | 600.9 | 625.5 | 606.4 | 608.0 |
| Mut 2% | 571.6 | 546.9 | 551.9 | 568.7 | 539.5 |
| Mut 4% | 391.6 | 406.9 | 401.9 | 408.7 | 409.5 |

Table 1. Performance of the best agents evolved in the fixed environmental condition obtained by systematically varying the number of evaluation trials, the mutation rate, and the level of stochasticity. Each number indicates the average results of 10 replications. The evolutionary process was continued for 50 millions evaluations. Data obtained by post-evaluating evolved agents for 1000 trials.

| 1 Trial | Stoch 0% | Stoch 10% | Stoch 20% | Stoch 30% | Stoch 40% |
|---|---|---|---|---|---|
| Mut 0.5 | 91.9 | 108.6 | 102.9 | 105.8 | 90.7 |
| Mut 1% | 119.9 | 130.2 | 118.9 | 111.8 | 103.7 |
| Mut 2% | 103.9 | 113.2 | 109.9 | 101.8 | 92.7 |
| 15 Trials | Stoch 0% | Stoch 10% | Stoch 20% | Stoch 30% | Stoch 40% |
| Mut 0.5 | 581.9 | 589.2 | 592.9 | 595.8 | 590.7 |
| Mut 1% | 669.9 | 670.2 | 678.9 | 681.8 | 673.7 |
| Mut 2% | 623.9 | 643.2 | 639.9 | 661.8 | 662.7 |
| 25 Trials | Stoch 0% | Stoch 10% | Stoch 20% | Stoch 30% | Stoch 40% |
| Mut 0.5 | 594.6 | 606.3 | 608.8 | 601.6 | 588.2 |
| Mut 1% | 696.9 | 707.6 | **709.5** | 700.5 | 698.5 |
| Mut 2% | 631.9 | 656.3 | 658.8 | 681.6 | 668.2 |
| 50 Trials | Stoch 0% | Stoch 10% | Stoch 20% | Stoch 30% | Stoch 40% |
| Mut 0.5 | 570.4 | 575.7 | 598.8 | 591.6 | 588.2 |
| Mut 1% | 665.6 | 675.6 | 698.9 | 697.8 | 675.7 |
| **Mut 2%** | 593.5 | 624.7 | 628.8 | 627.3 | 626.2 |
| **200 Trials** | **Stoch 0%** | **Stoch 10%** | **Stoch 20%** | **Stoch 30%** | **Stoch 40%** |
| **Mut 0.5** | 521.2 | 525.3 | 535.3 | 530.5 | 529.3 |
| **Mut 1%** | 604.2 | 615.2 | 617.2 | 613.5 | 611.6 |
| **Mut 2%** | 503.4 | 504.2 | 514.5 | 517.7 | 513.1 |

Table 2. Performance of the best agents evolved in the always-varying environmental condition obtained by systematically varying the number of evaluation trials, the mutation rate, and the level of stochasticity. Each number indicates the average results of 10 replications. The evolutionary process was continued for 50 millions evaluations. Data obtained by post-evaluating evolved agents for 1000 trials.

| 25 Trials | Stoch 0% | Stoch 10% | Stoch 20% | Stoch 30% | Stoch 40% |
|---|---|---|---|---|---|
| **Mut 1%** | 760.2000 | 755.0000 | 731.3000 | 739.5000 | 725.5000 |
| **Mut 2%** | 714.4000 | 704.2000 | 680.0000 | 685.9000 | 673.9000 |
| **Mut 4%** | 582.4000 | 578.4000 | 571.0000 | 582.8000 | 581.5000 |
| **50 Trials** | **Stoch 0%** | **Stoch 10%** | **Stoch 20%** | **Stoch 30%** | **Stoch 40%** |
| **Mut 1%** | 745.2000 | 732.9000 | 741.6000 | 739.8000 | 726.8000 |
| **Mut 2%** | 696.3000 | 702.7000 | 694.8000 | 675.4000 | 668.3000 |
| **Mut 4%** | 546.7000 | 543.0000 | 547.0000 | 525.9000 | 527.7000 |
| **100 Trials** | **Stoch 0%** | **Stoch 10%** | **Stoch 20%** | **Stoch 30%** | **Stoch 40%** |
| **Mut 1%** | 712.5000 | 714.8000 | 718.9000 | 686.4000 | 682.4000 |
| **Mut 2%** | 636.0000 | 643.5000 | 624.4000 | 658.9000 | 650.8000 |
| **Mut 4%** | 520.1000 | 495.1000 | 495.1000 | 464.3000 | 458.1000 |
| **150 Trials** | **Stoch 0%** | **Stoch 10%** | **Stoch 20%** | **Stoch 30%** | **Stoch 40%** |
| **Mut 1%** | 716.4000 | 694.2000 | 687.6000 | 694.8000 | 691.5000 |
| **Mut 2%** | 627.0000 | 634.7000 | 602.6000 | 654.8000 | 652.5000 |
| **Mut 4%** | 480.5000 | 471.8000 | 435.8000 | 458.3000 | 455.5000 |
| **200 Trials** | **Stoch 0%** | **Stoch 10%** | **Stoch 20%** | **Stoch 30%** | **Stoch 40%** |
| **Mut 1%** | 667.4000 | 676.2000 | 678.6000 | 663.8000 | 652.5000 |
| **Mut 2%** | 601.0000 | 603.8000 | 598.4000 | 584.8000 | 572.4000 |
| **Mut 4%** | 460.5000 | 451.8000 | 425.8000 | 448.2000 | 435.6000 |

Table 3. Performance of the best agents evolved in the intermediate environmental condition obtained by systematically varying the number of evaluation trials, the mutation rate, and the level of stochasticity. Each number indicates the average results of 10 replications of each experiment. Data obtained by post-evaluating evolved agents for 1000 trials.

## 6. Evaluating agents in systematically varied environmental conditions

To analyze how the behavior of the agents changes in systematically varied environmental conditions that covered the entire spectrum of possible variations we evaluated the agents for $3^6 = 729$ trials during which the state of the six variables that encode the characteristics of the plane and the most important initial characteristics of the cart were varied systematically. More specifically in each trial we selected one of the possible combination of the following six variables that assumed one of the three states indicated in square brackets: $\alpha$ [-0.1385, 0.0, 0.1385], $\mu_c$ [-0.15, 0.0, 0.15], $x$ [-0.75, 0.0, 0.75], $\dot{x}$ [-0.6, 0.0, 0,6], $\theta_1$ [-0.05235, 0.0, 0.05235], $\dot{\theta}_1$ [-0.0675, 0.0, 0.0675]. The angular position and velocity of the second pole ($\theta_2 \wedge \dot{\theta}_2$ were always set to 0.0.

**Acknowledgment**

Work partially funded by CAPES through the Brazilian program science without borders.